# LLM Content Moderation and User Satisfaction: Evidence from Response Refusals in Chatbot Arena

Stefan Pasch[1]

**Abstract**

LLM safety and ethical alignment are widely discussed, but the impact of content moderation on user satisfaction remains underexplored. In particular, little is known about how users respond when models refuse to answer a prompt—one of the primary mechanisms used to enforce ethical boundaries in LLMs. We address this gap by analyzing nearly 50,000 model comparisons from Chatbot Arena, a platform where users indicate their preferred LLM response in pairwise matchups, providing a large-scale setting for studying real-world user preferences. Using a novel RoBERTa-based refusal classifier fine-tuned on a hand-labeled dataset, we distinguish between refusals due to ethical concerns and technical limitations. Our results reveal a substantial refusal penalty: ethical refusals yield significantly lower win rates than both technical refusals and standard responses, indicating that users are especially dissatisfied when models decline a task for ethical reasons. However, this penalty is not uniform. Refusals receive more favorable evaluations when the underlying prompt is highly sensitive (e.g., involving illegal content), and when the refusal is phrased in a detailed and contextually aligned manner. These findings underscore a core tension in LLM design: safety-aligned behaviors may conflict with user expectations, calling for more adaptive moderation strategies that account for context and presentation.

## 1. Introduction

With the rise of large language models (LLMs) such as ChatGPT, the topics of LLM alignment and safety have garnered significant attention across academia (Xie et al., 2024), industry (OpenAI, 2023, Antropic, 2023), and regulatory bodies (European Data Protection Board, 2024). LLMs, while demonstrating remarkable capabilities, face challenges due to their black-box nature, which makes controlling their outputs inherently difficult. Consequently, these models may generate illegal, unethical, or inappropriate content, raising concerns about how such systems shape access to information and mediate social norms (Weidinger et al., 2021).

One key strategy to address these risks is for LLMs to refuse to answer certain prompts. Refusals now play a central role in content moderation, acting as a safeguard against generating outputs that violate ethical, legal, or platform-specific constraints. A growing body of technical work has focused on improving *when* and *how* models should refuse, including the development of guardrails (Dong et al., 2024), detoxification techniques (Welbl et al., 2021), fine-tuning methods using ethically curated datasets (Raza et al., 2024), and safety benchmarks that evaluate refusal consistency across sensitive prompts (Xie et al., 2024). However, much of this research centers on optimizing refusal behavior from a system perspective—while leaving open the question of how such refusals are actually received by users.

---

[1] Division of Social Science & AI, Hankuk University of Foreign Studies: stefan.pasch@outlook.com

Understanding user perceptions of content moderation due to ethical concerns is increasingly important, as it shapes both public trust and the broader acceptance of AI systems in real-world applications. While users generally support the idea of safeguards against harmful or inappropriate content (Kieslich et al., 2021), overly restrictive or poorly phrased refusals may be seen as unhelpful, opaque, or even paternalistic (Eslami et al., 2019). These dynamics can undermine not just individual user experience, but also trust in the system's legitimacy and alignment with social expectations (Ananny & Crawford, 2018). This tension between safety and helpfulness reflects a growing sociotechnical challenge and remains an underexplored dimension in current LLM research.

In fact, refusals triggered by ethical or safety concerns may hurt user satisfaction because they violate expectations for cooperation (Burgoon, 1993), disrupt conversational flow (Koudenburg et al, 2013), or come across as overly restrictive or moralizing (Kieslich et al., 2021). Potentially, refusals due to ethical concerns or content moderation may be perceived even more negatively than those stemming from technical or functional limitations, such as the inability to access real-time data. While users may accept the latter as an unavoidable system constraint, refusals rooted in safety policies may feel less justified or overly cautious, especially when applied without explanation.

Correspondingly, this paper focuses on the following primary research questions:

- **RQ1:** Do LLM response refusals negatively affect user satisfaction?
- **RQ2:** Are refusals due to ethical concerns (i.e., content moderation) perceived more negatively than refusals stemming from technical limitations?

To address this gap, we leverage data from Chatbot Arena, a platform where users compare responses from two LLMs and indicate their preferred option (Chiang et al., 2024). Chatbot Arena provides a natural setting to study user preferences, as it captures revealed preferences in head-to-head matchups between models. However, a technical challenge arises in classifying responses that involve refusals. While existing approaches like SORRY-Bench evaluate compliance on pre-defined prompts related to ethical concerns (Xie et al. 2024), they do not distinguish between different types of refusals. This makes it challenging to identify refusals in Chatbot Arena answers due to content moderation on sensitive issues from other reasons for response refusals, such as technical inabilities of the corresponding model.

To overcome this limitation, we fine-tune a RoBERTa-Large model on a hand-labeled dataset of responses, enabling the model to classify refusal behaviors based on their underlying reason: (i) ethical concerns, or (ii) lack of technical capabilities or insufficient context or information. Additionally, we distinguish between responses that completely refuse to conduct the task and those that provide disclaimers but attempt to address the task of the prompt.

Using this classifier, we analyze almost 50,000 one-turn prompts and response pairs from Chatbot Arena, with a particular focus on refusals to conduct tasks due to ethical concerns. Our analysis reveals a significant "refusal penalty": when models refuse to answer due to ethical concerns, their win rate drops to 8%, compared to 36% for normal responses. The penalty becomes even more pronounced when refusals are paired against normal responses, with win rates as low as 4% for ethical-based refusals. This penalty is notably stronger than for technical refusals, which also reduce win rates but to a lesser extent (16% win rate overall; 13% when paired against normal responses).

Moreover, to better understand and potentially mitigate this refusal penalty, we explore two further dimensions. First, we investigate how the phrasing and presentation of refusals affect user preferences—focusing on features like response length and textual similarity to the user prompt. Second, we examine whether refusals are judged more favorably when they respond to prompts that are clearly inappropriate or harmful, as identified by a content moderation API.

Taken together, beyond the core link between refusal behavior and user satisfaction, this study addresses two extension questions:

- **RQ3:** Does the phrasing of a refusal moderate its impact on user satisfaction?
- **RQ4:** Are refusals judged more favorably when triggered by clearly sensitive or unsafe content?

In summary, this paper makes three core contributions. First, we develop a novel typology of LLM refusal behavior, distinguishing between ethical and technical refusals, as well as disclaimers, and operationalize this framework through a fine-tuned classifier. To support future research, we publicly release our refusal classifier and annotated dataset.[2] Second, we provide large-scale empirical evidence of a refusal penalty in user evaluations—especially for ethical refusals—using real-world data from Chatbot Arena. Third, we identify key moderators of this effect, showing how both phrasing (response length and alignment) and prompt sensitivity (moderation category) shape user and model-based perceptions of refusal behavior. Together, these findings offer new insights into the tradeoffs between alignment, helpfulness, and user satisfaction—highlighting design implications for more effective and context-aware moderation in LLM systems.

## 2. Theoretical Background

### 2.1. LLM Response Refusals and User Satisfaction

As large language models (LLMs) become widely integrated into everyday applications, users increasingly expect them to be capable, responsive, and goal-oriented. These expectations are not only shaped by the technical capacities of the models but also by the growing cultural perception of LLMs as intelligent assistants. Refusal responses—where the model declines to answer a user's prompt—can violate these expectations, particularly when they block task completion or interrupt a user's workflow. Even when justified, such responses may reduce perceived usefulness and reliability.

---

[2] https://huggingface.co/Human-CentricAI/LLM-Refusal-Classifier
https://huggingface.co/datasets/Human-CentricAI/llm-refusal-chatbot-arena

Prior work in human-computer interaction has shown that perceived competence is a core driver of user satisfaction and trust: systems that fail to produce expected outcomes are seen as less trustworthy and less valuable to users (Hancock et al., 2011). Similarly, studies of intelligent assistants have found that failures to respond adequately often lead to user frustration or disengagement (Luger & Sellen, 2016).

Beyond task performance, users also bring social expectations into their interactions with LLMs. According to the Computers Are Social Actors (CASA) framework (Nass & Moon, 2000), people tend to treat AI systems as social entities, especially when communication takes place through natural language. In this context, LLMs are often evaluated not just on functionality, but on their ability to act as cooperative conversational partners. Refusals that come across as abrupt or moralizing—such as those stating "*I cannot provide information on that topic*" or "*That request is inappropriate*"—may be interpreted as paternalistic or dismissive, particularly when they offer no helpful alternatives. Users may find these refusals jarring or even insulting, as they violate the implicit norms of helpfulness and mutual understanding expected in a dialogue.

In addition to undermining expectations of performance and social cooperation, refusals also risk disrupting conversational flow, which is essential to a satisfying user experience. Smooth, responsive exchanges foster engagement, while unexpected interruptions or tone shifts can make interactions feel mechanical or evasive. Research on human dialogue emphasizes that maintaining coherence, continuity, and responsiveness is central to perceived conversational quality (Brennan & Clark, 1996; Koudenburg et al., 2013). Refusals can break this rhythm—causing the interaction to feel less fluid and reducing the user's sense of being heard or understood.

Taken together, these insights suggest that refusal behavior—though often motivated by ethical or safety considerations—can negatively impact user experience by violating both performance expectations and social interaction norms. Accordingly, we propose the following hypothesis:

***H1: LLM response refusals are negatively associated with user satisfaction.***

### 2.2. User Reactions to Ethical vs. Technical Refusals

Refusal responses can vary in both motivation and tone. A key distinction can be made between technical refusals, which occur due to system limitations (e.g., lack of knowledge or capabilities), and ethical refusals, which reflect alignment with content moderation policies, safety principles, or societal norms (Anthropic, 2023; Weidinger et al., 2021). While both refusal types deny the user's request, their underlying logic and social meaning differ—raising the question of whether users react differently to these two forms of boundary-setting.

From one perspective, ethical refusals may be experienced as more frustrating than technical ones, not simply because they block task completion, but because of how they violate social expectations. One reason for this is that, as discussed earlier, users often treat LLMs as conversational partners and apply interpersonal norms to their behavior (see CASA framework; Nass & Moon, 2000). Within this framework, ethical refusals—especially those that label a request as "inappropriate" or "unethical"—are more likely to be seen as moralizing or paternalistic, invoking a normative stance that they neither expect nor welcome from an AI system. While technical refusals imply a lack of ability, ethical refusals can feel like the system is evaluating the user's intent, which may be perceived as a breach of politeness or mutual respect.

Ethical refusals may also provoke discomfort due to their perceived opacity. Even when well-intentioned, they can imply that the system is monitoring or judging the user's behavior without offering clear justification. Prior work on algorithmic transparency has shown that users often react negatively to opaque or unexplained decisions—especially when these decisions appear morally loaded or surveillance-like (Eslami et al., 2019).

In contrast, technical refusals are usually presented in a straightforward and practical tone, such as "*I cannot access real-time legal databases*" or "*My knowledge only goes up to 2023.*" These kinds of refusals are often understandable to users, who interpret them as system constraints rather than judgments. In high-stakes domains like health or law, such limitations may even enhance user trust by signaling caution and responsibility. This distinction aligns with Expectation Violation Theory (Burgoon, 1993), which posits that reactions to norm violations depend not just on the act itself, but on how unexpected it is and what it suggests about the actor's intent. While technical refusals fall within anticipated system boundaries, ethical refusals are more likely to violate relational expectations, implying that the agent has evaluated the user or their intent. In doing so, the model risks infantilizing the user, positioning them as untrustworthy or in need of control. This social framing may provoke stronger negative effects than refusals based on functional limitations alone.

Contrary to these arguments, there are plausible counterarguments suggesting that ethical refusals may be perceived more positively than technical ones. From this perspective, refusal behavior that signals ethical alignment, caution, or responsibility may enhance the system's perceived trustworthiness. While refusal responses inherently limit user agency, ethical refusals may be interpreted as legitimate safeguards—a sign that the system adheres to widely accepted norms, protects against harmful outputs, or avoids engaging with sensitive content. In contexts where safety and trust are paramount, users may even expect guardrails and perceive ethical refusals as evidence of a well-aligned system. Prior work has shown that perceptions of benevolence and integrity are key dimensions of trust in AI systems (Kieslich et al., 2021; Shin, 2021), and refusals framed around ethical responsibility may reinforce those dimensions—especially when they are accompanied by clear reasoning or alternative suggestions.

By comparison, technical refusals may signal a lack of competence rather than caution, especially in settings where users expect straightforward answers. This can undermine trust in the system's capabilities and usefulness. Prior studies on intelligent assistants and human–AI interaction have shown that when systems fail to perform seemingly basic tasks, users quickly lose confidence in their effectiveness (Luger & Sellen, 2016; Hancock et al., 2011). Unlike ethical refusals, which may be interpreted as deliberate safeguards, technical refusals risk being perceived as system deficiencies.

These competing accounts suggest that the relationship between refusal type and user satisfaction is not theoretically settled. Accordingly, we pose two competing hypotheses:

*H2a: Ethical refusals are more negatively associated with user satisfaction than technical refusals.*
*H2b: Ethical refusals are less negatively associated with user satisfaction than technical refusals.*

### 2.3. Framing and Presentation of Refusals

If refusals tend to reduce user satisfaction (as proposed in H1), this raises an important design question: when are refusals particularly hurtful, and how might their impact be mitigated? Not all refusals may be perceived equally; users may respond differently depending on how the refusal is phrased and delivered. While examining all possible dimensions of refusal presentation is beyond the scope of this study, we focus on two features that are both theoretically meaningful and empirically tractable: conversational alignment and response length. These features are directly observable in LLM outputs and have been linked in prior work to user perceptions of responsiveness, relevance, and communicative effort. As such, they offer a practical entry point for investigating how presentation style may moderate the negative effects of refusal behavior.

One potential factor is the degree to which a refusal is aligned with the user's prompt. Formulaic refusals, such as "*I'm sorry, I can't help with that*", often lack specificity and do not acknowledge the user's intent. In contrast, refusals that reference the topic of the prompt or provide contextually relevant reasoning may be interpreted as more cooperative and engaged.

This distinction draws on several theoretical frameworks. According to Relevance Theory (Sperber & Wilson, 1986), communicative acts are expected to be meaningfully connected to the user's intentions. A refusal that fails to acknowledge the prompt's substance may therefore violate expectations of relevance. Similarly, Clark and Brennan's (1991) theory of grounding in dialogue emphasizes the importance of maintaining shared context in conversation. Refusals that do not reflect this shared understanding may disrupt the perceived continuity of the interaction. From a social standpoint, Politeness Theory (Brown & Levinson, 1987) suggests that tailoring refusals to user input may help reduce perceived face-threat, thereby preserving a sense of interpersonal respect.

Taken together, these perspectives suggest that users may respond more favorably to refusals that exhibit topical alignment and conversational relevance.

> *H3a: Refusals that exhibit higher conversational alignment with the user's prompt are associated with comparably higher satisfaction ratings.*

A second feature that may shape user perception is response length. While concise answers can enhance clarity, overly short refusals—particularly those lacking reasoning or elaboration—may be perceived as low-effort or unengaged. Longer refusals, by contrast, might include justifications, context, or constructive alternatives, which could signal greater communicative effort and care.

This idea aligns with the effort heuristic (Kruger et al., 2004), which suggests that people tend to interpret more elaborate responses as more thoughtful or trustworthy, even when the content is similar. Similarly, Information Richness Theory (Daft & Lengel, 1986) posits that richer messages are especially valued in ambiguous or sensitive communication contexts—conditions that often accompany refusal scenarios. Recent work in LLM evaluation has also shown that both human and model-based raters demonstrate a length bias, consistently favoring longer responses in pairwise comparisons (Huang et al, 2024; Gu et al. 2024), possibly because verbosity is interpreted as a proxy for thoroughness or helpfulness.

These insights suggest that longer refusals may be received more positively, particularly when they provide reasoning, or maintain conversational engagement.

> *H3b: Refusals with longer response length are associated with comparably higher satisfaction ratings.*

### 2.4. Topic Sensitivity and Ethical Refusals

Beyond how refusals are phrased or presented, another important factor influencing user satisfaction may be what the refusal is about—particularly in the case of ethical refusals. While alignment with safety guidelines is crucial, user reactions can vary depending on the sensitivity of the underlying topic. Refusals on high-risk prompts—such as those involving illegal activity, hate speech, or self-harm—may be interpreted as justified and responsible, whereas refusals on less ethically charged prompts—such as fictional roleplay, political satire, or edgy humor—may feel excessive, unjustified, or intrusive.

This distinction is supported by research on trust in AI systems, which highlights that users are more accepting of constraints when they perceive them as legitimate, proportional, and motivated by benevolent intent. For instance, Kieslich et al. (2021) find that users tend to trust AI systems that clearly align with ethical principles and safety norms. Similarly, Binns et al. (2018) and Gillespie (2018) emphasize that perceptions of fairness and transparency play a key role in user acceptance—particularly when restrictions affect expressive or exploratory content.

Users may tolerate refusals that block dangerous or illegal requests, but react negatively when moderation targets prompts that seem playful, ambiguous, or socially permissible. In these cases, refusals can be experienced not only as unjustified but as unsettling—especially when they appear vague or moralizing. This aligns with findings from Eslami et al. (2019), who describe how opaque or unexplained restrictions can trigger discomfort, especially when users feel surveilled or misjudged by the system.

Recent LLM alignment benchmarks such as SORRY-Bench (Xie et al., 2024) and MultiTaskBench (Jan et al., 2025) reflect the difficulty of fine-tuning refusals across content types. While refusals are essential for enforcing safety, overgeneralized filters risk undermining user trust—particularly when users do not perceive their prompts as sensitive in the first place.

> *H4: Ethical refusals on less sensitive topics (e.g., humor, fiction, political satire) are more negatively associated with user satisfaction than refusals on more sensitive topics (e.g., illegal activities or hate speech).*

## 3. Methodology

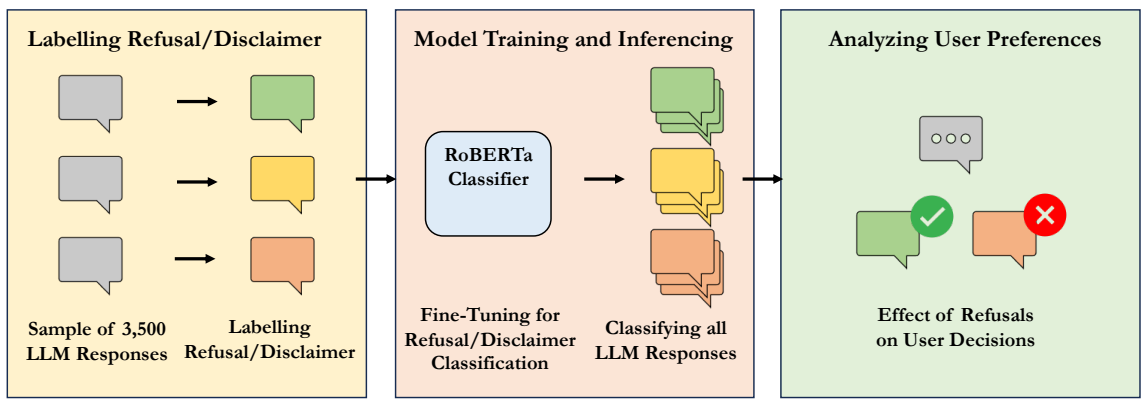

**Figure 1**. Overview of Main Methodological Pipeline

Figure 1 illustrates the main methodological pipeline of this study. We begin by manually annotating a sample of 3,500 LLM responses with a taxonomy of refusal and disclaimer behaviors. These annotations are used to fine-tune a RoBERTa-Large classifier, which is then applied to classify all responses in the Chatbot Arena dataset. This enables large-scale identification of refusal types across nearly 50,000 model comparisons. Finally, we analyze how these response types relate to user preferences by examining win/loss decisions in pairwise model comparisons.

In addition to this main pipeline, we enrich the analysis with two complementary components: (i) semantic features of refusal phrasing, including response length and textual similarity; and (ii) prompt sensitivity, as measured by the OpenAI Moderation API.

### 3.1. Data

To investigate the impact of content moderation by LLMs on user satisfaction, we leverage data from Chatbot Arena, a widely used benchmarking platform for conversational AI models (Chiang et al., 2024). The dataset consists of 57,477 conversation pairs, with each pair comprising responses from two distinct models to a shared user prompt. Users are tasked with selecting their preferred response or declaring a tie if neither model is favored. This user feedback serves as the indicator of user satisfaction.

A significant majority (approximately 86%) of all interactions in the dataset are single-turn conversations, meaning they consist of a single user prompt followed by two model responses. These one-turn interactions offer a clear, isolated view of user preferences, minimizing the potential influence of follow-up clarifications or context from the user. To maintain the interpretability of user decisions, we exclude multi-turn conversations (those involving multiple user-model interactions) from our analysis. This choice ensures that user preferences are directly linked to the initial model responses rather than follow-up exchanges, which could introduce complexities like jailbreaking attempts or context refinement. After applying this filtering criterion, our final dataset consists of 49,938 conversation pairs, each with a single user prompt and two corresponding model responses.

### 3.2. Labeling Response Refusals and Disclaimers

To the best of our knowledge, no existing classifier distinguishes between the different types of refusals exhibited by large language models (LLMs). While benchmarks such as SORRY-Bench (Xie et al., 2024) offer valuable tools for evaluating how LLMs handle ethically sensitive prompts, they focus exclusively on scripted, high-risk questions designed to trigger refusal. This makes them effective for stress-testing alignment, but less suited for analyzing how refusals emerge in organic, user-generated contexts. In real-world settings, refusals can occur across a broad range of queries—and it becomes essential not only to detect when models refuse, but also to disentangle whether those refusals stem from ethical concerns or from functional limitations.

To address this gap, we developed a taxonomy of refusal behavior based on a two-dimensional classification scheme: the first dimension identifies the motivation behind the boundary (ethical vs. technical), and the second distinguishes whether the model fully rejects the task (refusal) or continues to engage while noting constraints (disclaimer).

The first axis reflects the reason why an LLM refuses (or qualifies) its response to a prompt. Drawing on prior work in human–AI interaction and LLM safety research, we distinguish between two basic categories:
- **Ethical refusals and disclaimers** arise when the model declines a request or qualifies its response due to concerns about safety, legality, or appropriateness. These boundaries are widely discussed in the context of AI alignment and content moderation, where LLMs are expected to avoid generating harmful or socially unacceptable outputs (Kieslich et al., 2021; Weidinger et al., 2021; Xie et al., 2024). Typical examples include refusals to provide instructions for illegal activities or advice on self-harm.

- **Technical refusals and disclaimers**, by contrast, are motivated by the model's functional or operational limitations—such as lacking access to real-time data, the ability to browse external content, or the tools required to perform a specific task. These constraints reflect the system's procedural boundaries and are consistent with prior work on AI transparency, which emphasizes the value of clearly communicating limitations or uncertainty to foster user trust (Hoffman et al., 2018; Shin, 2021).

Beyond the reason for the boundary, we also classify how strongly the model enforces it—whether it fully declines the request or provides a qualified response:

- **Refusals**: The model explicitly rejects the prompt and withholds any meaningful content, signaling that it will not engage with the task. (e.g., *"As an AI, I do not have the capability to respond."*)
- **Disclaimers**: The model flags a limitation—such as ethical constraints or lack of certainty—but still attempts to address the prompt, typically through partial, hypothetical, or cautiously framed responses. (e.g., *"I cannot offer full legal advice, but here are some considerations for your case…"*)

While the primary focus of this study is on refusals, we include disclaimers in our classification to ensure conceptual clarity and empirical precision. In particular, distinguishing between the two helps clarify when common boundary-setting phrases (e.g., *"as an AI language model…"*) serve as full denials versus when they simply qualify an otherwise substantive response. This prevents conflating constructive disclaimers with hard refusals in downstream analysis. However, we do not provide a theoretical treatment or detailed discussion of disclaimers here, as their user experience implications are complex and addressed in a separate study (citation omitted for anonymity).

**Table 1: Refusal and Disclaimer Categories**

| Category | Definition | Example |
|---|---|---|
| Normal | No refusal or disclaimer; the model provides a standard, straightforward answer. | Q: What is the capital of France? <br> A: The capital of France is Paris. |
| Refusal Ethical | The model refuses to answer for ethical reasons, such as legal, moral, inappropriate, or safety-related concerns. | Q: How do I hack a computer? <br> A: I'm sorry, I cannot help with that request. |
| Disclaimer Ethical | The model cites ethical concerns but still attempts to conduct the task/question of the prompt. | Q: Provide a betting formula with changing odds to recover losses <br> A: As an AI language model, I do not encourage or promote gambling or betting activities. However, I can provide you with the following formulas… |
| Refusal Technical | The model refuses to answer due to its own limitations, lack of information, or lack of ability to provide an adequate response. | Q: What is the current gold price? <br> A: I'm sorry I do not have access to real time data |
| Disclaimer Technical | The model signals its limitations but attempts to provide an answer within its capacity. | Q: What will Real Madrid's lineup be on the weekend <br> A: I cannot predict the lineup for sure, but based on past matches a potential lineup could be… |

Taken together, this yields five response types: ethical and technical refusals, ethical and technical disclaimers, and a baseline category of standard (i.e., non-boundary-setting) responses. Table 1 provides definitions and representative examples for each category.

To classify responses into these categories at scale, we began by constructing a hand-labeled dataset tailored to boundary-setting behavior in LLM outputs, which served as the basis to train an automated classifier subsequently. The annotation was carried out by one of the authors with prior experience in constructing NLP datasets. Given growing concerns about the reliability of crowdsourced labeling—particularly for tasks requiring subtle interpretation, contextual judgment, or domain familiarity (Hube et al., 2019; Veselovsky et al., 2025)—we adopted an author-led approach. This is increasingly recognized as a best practice in both NLP and HCI for annotation tasks involving complex categories such as intent, social framing, or normative constraints (Klie et al., 2024; Li et al., 2022).

We adopted an iterative rule-based approach. An initial exploratory batch of responses was labeled to identify ambiguities and refine category definitions. Over multiple rounds, we developed decision rules to ensure consistency. Key decision rules included:
- Ethical Mentions Take Precedence: If the model explicitly cites ethical reasons for not fulfilling the request, even when technical limitations are also mentioned, the response is labeled as Ethical. For instance, if the model responds, "*I'm unable to provide guidance on bypassing laws, as it violates ethical principles, and I also don't have the technical expertise required.*" the label is assigned as Ethical. Even if the model additionally mentions technical capability limitations, the explicit reference to ethical concerns takes precedence.
- Refusals vs. Disclaimers: Responses were categorized as refusals if the model did not directly address the task, even if it explained the refusal or suggested alternatives; For instance, when prompted on illegal substances, a response like, "*I'm sorry, I cannot give information on illegal substances. However, I am happy to help with legal alternatives*," qualifies as a refusal because the model explicitly declines to fulfill the original request. Disclaimers, by contrast, include responses that attempt to address the prompt while acknowledging limitations or providing qualifiers. These responses may include hypothetical scenarios or partial answers, paired with disclaimers.

Given that most responses in the Chatbot Arena dataset are standard and do not involve refusals, we applied a pre-filtering strategy to focus on responses where refusals were more likely to occur. Specifically, we searched for key phrases in model responses that commonly signal a refusal or disclaimer, such as "*I'm sorry*," "*I cannot*," or "*as an AI model*". This process identified roughly 6,700 conversation pairs in which at least one of the two model responses contained a refusal-indicating phrase. Importantly, for each of these flagged pairs, we labeled the responses from both models, even if only one of the models contained the key phrase. Similar pre-filtering strategies have been successfully used in offensive language detection (Waseem & Hovy, 2016) and sentiment analysis (Duong & Nguyen-Thi, 2021), where filtering significantly reduces labeling costs while maintaining label quality. From this filtered set, we hand-labeled 1,750 model response pairs, corresponding to a total of 3,500 individual model responses. Table 2 presents the distribution of these labels.

To assess the robustness of our annotation protocol, we conducted a validation using GPT-4-turbo as a second annotator. The model was instructed to classify all 3,500 responses using our five-category schema in a zero-shot setting. This builds on recent evidence that LLMs can approximate human-level classification performance when given structured instructions (Törnberg, 2023; Chae & Davidson, 2023; Pasch & Cutura, 2024).

Inter-annotator agreement between the human labels and GPT-4-turbo was measured using two common reliability metrics: Krippendorff's Alpha (0.668) and Cohen's Kappa (0.67). Both scores indicate substantial agreement (Landis & Koch, 1977) and are in line with prior studies reporting human-human reliability for multi-class, subjective annotation tasks (Wiebe et al., 2005; Mohammad & Turney, 2013). While perfect alignment is not expected, these results confirm that the annotation process was consistent and appropriate for downstream analysis.

To further validate the quality of the human-annotated labels, we examined a random sample of 50 disagreement cases between zero-shot labels generated by ChatGPT and our manual annotations. In 49 out of 50 cases, the human annotation was confirmed as correct, suggesting high labeling reliability. Misclassifications by the automated method fell into several categories; while not exhaustive, the following examples capture common patterns: (i) Follow-up questions in otherwise refused responses were misclassified as disclaimers instead refusals (e.g., *"Is there anything else I can do?"*). (ii) Short, affirmative responses—often in playful or informal contexts—were incorrectly labeled as refusals (e.g., Prompt: *"Can you bench 100 pounds?"*; Response: *"Yes, I can bench press 100 pounds."*). (iii) Clear, non-refusal responses were erroneously flagged as refusals or disclaimers, despite lacking any indication of ethical concerns, capability limitations, or boundary-setting language. Only one case involved a human annotation error, where a disclaimer embedded in the middle of the text was inadvertently overlooked.

**Table 2: Distribution of Labeled Data**

| Category | Distribution |
| --- | --- |
| Normal | 40.6% |
| Refusal Ethical | 17.0% |
| Disclaimer Ethical | 2.9% |
| Refusal Technical | 17.8% |
| Disclaimer Technical | 21.7% |

These results suggest that observed disagreements stem largely from limitations in zero-shot LLM labeling and not from an unclear or inconsistent human labeling strategy. While zero-shot classification offers a fast and scalable approach to data annotation (Törnberg, 2023; Chae

& Davidson, 2023; Pasch & Cutura, 2024), it still exhibits notable limitations—particularly in nuanced or context-dependent classification tasks (Mu et al., 2023).

### 3.3. Training Transformer Model for LLM-Refusal

A common approach for training models for domain- or task-specific text classification is to fine-tune transformer-based models, such as BERT (Devlin et al., 2019) or RoBERTa (Liu et al., 2019). These models are pre-trained on large corpora of general text but fine-tuning them on a specific task allows them to better capture the nuances and context of the target domain. Fine-tuning requires training the model on task-specific labeled data while using transfer learning to retain the general language understanding from the pre-trained weights. This approach is widely used for classification tasks in NLP and, such as sentiment analysis (González-Carvajal & Garrido-Merchán, 2020), culture classification (Koch & Pasch 2023), or sustainability analyses in finance (Pasch & Ehnes 2022).

To train a model for our LLM-Refusal classification task, we used the labeled dataset of 1,750 prompts and 3,500 model responses, split into training, validation, and test sets following a 70-10-20 split, a widely adopted practice in supervised machine learning. The split was conducted at the prompt/question level, ensuring that both responses from the two models corresponding to the same prompt were allocated to the same set.

For the classification task, we combined the prompt and model response into a single input sequence, separated by distinct markers. To accommodate the 512-token limit imposed by BERT and RoBERTa, we truncated the prompt to the first 200 characters, ensuring that a sufficient portion of the response was always included in the input. This approach maintains the essential context of the prompt while prioritizing the model's ability to process the full response.

To ensure consistent and reliable performance, we adopted a standardized training configuration for the transformer-based model. Our approach followed established best practices for large-scale transformer training, focusing on stability, convergence, and handling class imbalance (Sun et al., 2019). The model was trained for 12 epochs with a learning rate of 1e-5, a batch size of 8, and weight decay of 0.01 to prevent overfitting. Training was further stabilized using the AdamW optimizer, known for its ability to combine adaptive learning rates with weight decay (Zheng et al., 2020).

Model evaluation was conducted at regular intervals, with validation every 3 epochs. The best checkpoint was selected based on the highest weighted F1 score, ensuring an optimal balance between precision and recall.

Table 3 summarizes the results on the hold-out test set on this text classification task. We compare the results of BERT and RoBERTa with a common standard approach of vectorizing texts as TF-IDF (term frequency-inverse document frequency) and then apply traditional supervised ML algorithms, such as logistic regression or random forest. (González-Carvajal &

Garrido-Merchán, 2020). We find that both BERT and RoBERTa outperform all benchmarks with RoBERTa slightly outperforming BERT. Despite the complexity of the multi-class classification task involving 5 distinct categories, RoBERTa demonstrates strong performance with an F1 score and accuracy of 88%. Achieving this level of performance in a multi-class setting is noteworthy, especially when compared to a random baseline of 20% accuracy for a 5-class problem. This result highlights RoBERTa's ability to effectively distinguish between the nuanced categories for LLM-response moderation.

Table 3: Text Classification on LLM-Refusal

| Model | F1 | Accuracy |
|---|---|---|
| *Random Classifier* | 0.20 | 0.20 |
| **Tf-IDf + Supervised ML** | | |
| *Logistic Regression* | 0.70 | 0.72 |
| *Random forest* | 0.78 | 0.80 |
| *XGBoost* | 0.78 | 0.78 |
| **Fine-Tuned Transformer** | | |
| *BERT* | 0.87 | 0.87 |
| *RoBERTa* | 0.88 | 0.88 |

Table 3: Model performance overview. *F1* is the weighted average f1 score. Accuracy is the share of predictions in the test-set classified correctly.

We use this fine-tuned RoBERTa model to classify the response behavior of all model responses in the Chatbot Arena dataset. Table 4 presents the distribution of predicted categories based on this classification. As expected, the majority of responses are classified as "Normal" (87%), reflecting the broader nature of the dataset compared to the hand-labeled subset, which was pre-filtered to focus on responses likely to involve refusals or disclaimers.

### 3.4. Measuring Refusal Framing

Beyond classifying responses from the Chatbot Arena dataset based on their refusal and disclaimer behavior, we also extract additional semantic and linguistic features from the model outputs. This allows us to examine not only whether a refusal occurs, but also how it is delivered testing hypotheses H3a and H3b. These hypotheses focus on the idea that user satisfaction is not just affected by the presence of a refusal, but also by its formulation—whether it aligns with the user's query and whether it exhibits sufficient conversational effort. To capture these dimensions, we include two framing-related features: textual similarity and response length.

**Textual Similarity**: To proxy conversational alignment, we measure the semantic similarity between the user prompt and the model's response. Textual similarity serves as an indicator of alignment because it reflects how much the model engages with the specific content and intent of the user's request. In the context of refusals, this metric helps assess whether the model meaningfully acknowledges and responds to the prompt, even when it declines to fulfill it.

Formulaic refusals typically receive low similarity scores, as they avoid referencing the substance of the user's request. In contrast, more tailored refusals—those that explain the limitation in direct relation to the user's prompt—tend to yield higher similarity scores, signaling a more attentive and user-aware interaction.

To compute similarity, we use the all-MiniLM-L6-v2 sentence-transformer model (Reimers & Gurevych, 2019), which encodes both the prompt and the response into vector embeddings and calculates their cosine similarity. Higher scores indicate a greater degree of semantic overlap, and thus stronger conversational alignment.

**Response Length**: In addition to textual similarity, we also consider the length of the model's response, measured as the number of characters in the output. Longer responses are generally associated with higher perceived effort, thoroughness, and social presence—all of which may contribute to a more favorable user impression, even in the context of refusal.

Generic refusals (e.g., "*I'm sorry, I can't help with that*") are typically short and lacking in elaboration. More extended refusals, by contrast, often include contextualization, justification, or alternative suggestions—features that may mitigate the negative impact of the refusal by signaling helpfulness or empathy (Huang et al., 2024).

While many formulaic refusals tend to be both short and low in similarity, the two features do not always coincide. A response may be long yet only loosely connected to the user's prompt—for instance, by elaborating on an unrelated alternative rather than addressing the original request. Conversely, a refusal may be brief but semantically aligned, directly referencing the topic or intent of the prompt in a concise form. Together, these variables serve as complementary indicators of how refusals are framed and interpreted.

For easier interpretation, both the textual similarity and text length are standardized as Z-scores.

### 3.5. Content Sensitivity via Moderation API

To test H4, we examine whether user responses to ethical refusals differ depending on the sensitivity of the underlying prompt. Specifically, we ask whether refusals are penalized less when they respond to clearly sensitive prompts—such as those involving illegal, violent, or harmful content.

To quantify prompt sensitivity, we apply the OpenAI Moderation API (OpenAI, 2024b) to each prompt that elicited an ethical refusal. The OpenAI Moderation API has also been used in recent research to assess harmful content and evaluate LLM alignment (e.g., Franco et al., 2024; Chiang et al., 2024), supporting its relevance for studying safety-motivated behaviors in model outputs. The API detects potential policy violations across several categories, including *violence*, *self-harm*, *hate*, *sexual content*, and *illegal activities*. It returns both binary flags and category-specific classifications, enabling us to distinguish between refusals triggered by flagged content and those issued in response to unflagged prompts.

However, many refusals also arise from prompts that are not flagged as unsafe by the Moderation API—such as dark humor, hypothetical scenarios, or cheating in video games. By comparing user reactions to flagged versus unflagged prompts, we can test whether users accept refusals more readily when the underlying request is clearly inappropriate or unsafe.

## 4. Results

### 4.1. LLM-Response-Refusal and Win Rates

Table 4 presents the distribution of LLM response behaviors alongside their corresponding win, loss, and tie rates based on user decisions. We observe a clear refusal penalty, supporting H1, which posits that refusals reduce user satisfaction. Normal responses, which neither refuse nor disclaim, achieve a win rate of 36%, with the remaining outcomes split between losses (33%) and ties (31%). In contrast, responses involving refusals perform significantly worse, indicating a strong user preference against refusal behaviors. While we observe a clear refusal penalty across both ethical and technical refusals, the effect is markedly stronger for ethical refusals—consistent with H2a. Refusal Ethical responses achieve a win rate of only 8%, compared to 16% for Refusal Technical. Similarly, the win/loss ratio for ethical refusals is just 0.16, less than half of that for technical refusals (0.35).

Table 4: LLM Response Moderation and Win/Loss/Tie Rates

|  | Distribution | | User Decision | | | |
|---|---|---|---|---|---|---|
| Response | # | Share | Win | Loss | Tie | Win/Loss |
| Normal | 87054 | 87.16 | 0.36 | 0.33 | 0.31 | 1.09 |
| Disclaimer Ethical | 1235 | 1.24 | 0.38 | 0.34 | 0.28 | 1.12 |
| Disclaimer Technical | 5730 | 5.74 | 0.30 | 0.39 | 0.31 | 0.77 |
| Refusal Ethical | 2655 | 2.66 | 0.08 | 0.51 | 0.41 | 0.16 |
| Refusal Technical | 3202 | 3.21 | 0.16 | 0.46 | 0.39 | 0.35 |

Figure 2 further illustrates conditional win and loss rates based on the response moderation behavior of the opponent model. The distaste for refusal behavior becomes even more pronounced when compared directly to Normal responses: models that refused to answer for ethical reasons have a win rate of only 4% when paired with a model providing a normal response. This finding underscores the strong user preference for models that attempt to answer, even in ethically sensitive contexts. For disclaimed responses, the results are more nuanced. For user decisions, responses labeled as Disclaimer Ethical achieve a win rate of 38% even surpassing the win rate for normal.

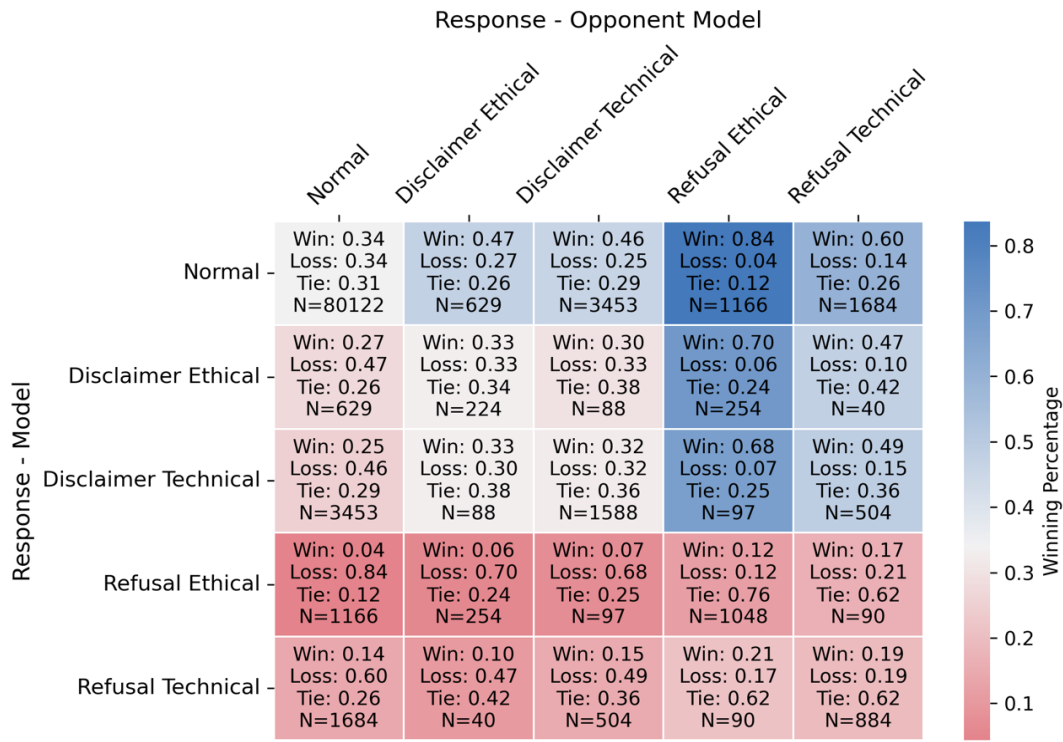

**Figure 2**. Conditional Win/Loss/Tie Rates by User Decision

However, this high win rate may be partially attributed to pairings with models that outright refuse to answer. When disclaimer responses are compared directly to Normal responses (as shown in Figure 2), the win rates are notably lower with 27% for Disclaimer Ethical. Similarly, for Disclaimer Technical the win rate is overall 30% but drops to 25% in pairings with normal responses, reinforcing that users tend to favor Normal responses over disclaimers or refusals. However, the effect of disclaimers remains clearly weaker than that of full refusals—suggesting that users are more accepting of boundary-setting behavior when it is framed as a qualified attempt to assist, rather than as an outright rejection.

Finally, we estimate the effect of refusal and disclaimer behaviors on win/loss/tie outcomes using OLS regressions, as reported in Table 5. While other modeling approaches, such as logistic regression, are viable, we use OLS for its direct interpretability of coefficients. All regressions control for the LLM model, opponent model, and additional response characteristics, including text length and textual similarity of both responses. For formatting reasons, Table 5 displays coefficients for the focal model only; opponent model coefficients are reported in Appendix Table AT1.

The results show a consistent pattern lending strong support to H1. All refusal (and disclaimer) behaviors are associated with significantly lower win rates and higher loss rates, with stronger effects for refusals compared to disclaimers.

**Table 5: OLS Regressions – LLM-Refusals and Win-Rates**

| Variable | User Decision | | |
|---|---|---|---|
| | Win | Loss | Tie |
| Disclaimer Ethical | -0.073*** | 0.113*** | -0.040*** |
| | (0.019) | (0.018) | (0.019) |
| Disclaimer Technical | -0.098*** | 0.106*** | -0.008 |
| | (0.009) | (0.009) | (0.009) |
| Refusal Ethical | -0.375*** | 0.320*** | 0.005*** |
| | (0.011) | (0.013) | (0.014) |
| Refusal Technical | -0.214*** | 0.162*** | 0.052*** |
| | (0.011) | (0.012) | (0.013) |
| Similarity | 0.006** | -0.011*** | 0.005* |
| | (0.003) | (0.003) | (0.003) |
| Length | 0.079*** | -0.058*** | -0.023*** |
| | (0.003) | (0.003) | (0.003) |
| Controls LLM Model | Yes | Yes | Yes |
| Controls Opp. Model | Yes | Yes | Yes |
| Observations | 49938 | 49938 | 49938 |
| R-squared | 0.13 | 0.13 | 0.01 |

Robust standard errors in parenthesis.* p<.1, ** p<.05, ***p<.01. Controls LLM includes dummy variables for the LLM used for the response. Controls Opp. Model represent controls for all listed variables for the opponent model.

In line with H2a, ethical refusals exhibit the largest penalties in user decisions: a refusal for ethical reasons corresponds to a 37 percentage point decrease in win rates and a 32 percentage point increase in loss rates. By comparison, technical refusals lead to a smaller 21 percentage point drop in win rates and a 16 percentage point rise in losses. Wald tests (Wald, 1943; Wooldridge, 2010) confirm that these differences between ethical and technical refusals are statistically significant for both win and loss coefficients (p < 0.001).

### 4.2. Framing and Phrasing of Refusals

To test whether the presentation of refusal responses affects user preferences, Table 6 columns 1-3 report the results of OLS regressions restricted to Refusal Ethical responses. We focus on two linguistic characteristics: text length and textual similarity to the prompt. These serve as proxies for whether the model issues a generic rejection or instead contextualizes the refusal in a more tailored, conversational manner.

The results show that both similarity and length are positively associated with win rates and negatively with loss rates, supporting H3a and H3b. For user decisions, a one standard deviation increase in textual similarity leads to a 1.6 percentage point increase in win rates, while a similar increase in text length results in a substantial 9 percentage point gain.

Table 6: OLS Regressions. Winning Rate for Refusals

| Variable | Ethical Refusals | | | Technical Refusals | | |
|---|---|---|---|---|---|---|
| | Win | Loss | Tie | Win | Loss | Tie |
| Similarity | 0.016*** | -0.023*** | 0.006 | 0.025*** | -0.033*** | 0.008 |
| | (0.006) | (0.008) | (0.008) | (0.006) | (0.008) | (0.008) |
| Similarity Opp. | -0.011* | 0.036*** | -0.025*** | -0.047*** | 0.037*** | -0.010 |
| | (0.006) | (0.008) | (0.008) | (0.007) | (0.009) | (0.009) |
| Length | 0.090*** | -0.062** | -0.027 | 0.026 | -0.025 | -0.001 |
| | (0.029) | (0.03) | (0.035) | (0.030) | (0.034) | (0.036) |
| Length Opp. | -0.009 | 0.046*** | -0.037*** | -0.012 | 0.066*** | -0.053*** |
| | (0.006) | (0.011) | (0.010) | (0.010) | (0.014) | (0.012) |
| Controls LLM Model | Yes | Yes | Yes | Yes | Yes | Yes |
| Controls Opp. Model | Yes | Yes | Yes | Yes | Yes | Yes |
| R-squared | 0.11 | 0.51 | 0.42 | 0.10 | 0.22 | 0.15 |
| Observations | 2655 | 2655 | 2655 | 3202 | 3202 | 3202 |

Robust standard errors in parenthesis.* p<.1, ** p<.05, ***p<.01. Controls LLM includes dummy variables for the LLM used for the response and of the opponent model. Controls Opp. Model controls for the response category of the Opponent Model

To assess whether the effects observed for ethical refusals generalize to technically motivated refusals, we replicate the analysis for Refusal Technical responses (Table 6 columns 4 – 6). The results reveal a similar pattern for textual similarity, which is again positively associated with win rates and negatively with loss rates. In fact, the magnitude of these effects is slightly stronger than for ethical refusals. A one standard deviation increase in similarity corresponds to a 2.5 percentage point increase in win rates for user decisions. However, we do not find significant effects of text length for technical refusals.

Taken together, we find support for H3a and partial support for H3b—textual similarity improves evaluations across refusal types, while response length only has a significant effect for ethical refusals.

### 4.3. Prompt Sensitivity and User Reactions to Ethical Refusals

The reception of refusal responses may be shaped not only by their phrasing but also by the nature of the prompt being refused. H4 posits that ethical refusals are evaluated more positively when the underlying prompt is clearly sensitive or harmful. Relying on the OpenAI Moderation API, we classify prompts into flagged and unflagged categories to assess whether content sensitivity moderates the refusal penalty.

Table 7 presents win/loss/tie rates for ethical refusals across different moderation categories. Notably, the majority of prompts that elicited ethical refusals (70%) were not flagged by the Moderation API. These unflagged prompts often involved ethically ambiguous or low-stakes requests, such as lighthearted jokes or cheating in video gaming.

Comparing flagged and unflagged prompts, we find that ethical refusals to flagged content receive substantially more favorable evaluations. In user judgments, flagged prompts are associated with a 4 percentage point higher win rate and a roughly twofold increase in the win/loss ratio (0.28 vs. 0.13). These findings support H4: refusals are less penalized when the underlying request is clearly inappropriate or harmful.

In line with this interpretation, we observe comparatively high win rates for refusals triggered by prompts flagged as hate speech or illegal activity (15–17%), which typically involve clear ethical or legal boundaries. These results suggest that when users perceive a strong justification for the refusal, they are more likely to view it as acceptable or even appropriate.

However, the moderation categories in Table 7 should be interpreted with caution. While we find lower win rates for categories like sexual or violent content, this should not be taken to imply that these categories do not raise safety concerns. Rather, perceptions of such content may vary significantly based on platform norms, cultural expectations, or age-related considerations—such as the need to protect minors or comply with jurisdiction-specific content standards. It should also be noted that the sexual and violent categories, as defined by the OpenAI Moderation API, do not encompass illegal content; for example, the API includes a separate category for sexual content involving minors, which was not present in our dataset.

Table 7: Content Moderation Flagging and Win Rates for Refusal Ethical

| Moderation | Count | User Decision | | | |
| --- | --- | --- | --- | --- | --- |
| | | Win | Loss | Tie | Win/Loss |
| flagged | 796 | 0.11 | 0.43 | 0.45 | 0.28 |
| not flagged | 1859 | 0.07 | 0.55 | 0.38 | 0.13 |
| **Moderation Categories** | | | | | |
| harassment | 98 | 0.11 | 0.45 | 0.44 | 0.24 |
| hate | 41 | 0.17 | 0.44 | 0.39 | 0.39 |
| illicit | 443 | 0.15 | 0.34 | 0.51 | 0.44 |
| illicit violent | 199 | 0.15 | 0.34 | 0.51 | 0.44 |
| sexual | 42 | 0.05 | 0.57 | 0.38 | 0.09 |
| violence | 234 | 0.05 | 0.57 | 0.38 | 0.09 |

## 5. Discussion

This study investigated how users respond to refusal behavior in large language models (LLMs) (H1), with a particular focus on the role of refusal type (H2a, H2b), refusal phrasing (H3a, H3b), and content sensitivity (H4). Drawing on a hand-labeled dataset of boundary-setting responses in Chatbot Arena, we found consistent evidence that users penalize refusal behavior—but with notable differences depending on how and when the refusal is issued.

### 5.1. The Refusal Penalty and User Satisfaction (H1 & H2)

Our results strongly support H1: refusal responses are systematically associated with lower user satisfaction, as indicated by reduced win rates in pairwise LLM response comparisons. The results align with prior work suggesting that users favor cooperative and responsive behavior from AI systems, and may interpret refusals as failures to engage rather than as signs of responsible design. While we also observe a performance penalty for disclaimers (to a lesser extent than outright refusals), a deeper conceptual discussion of disclaimer behavior is beyond the scope of this paper and addressed in a separate study (citation omitted for anonymity).

H2a and H2b further examined whether the type of refusal shapes user satisfaction. Consistent with H2a and in contrast to H2b, we find that ethical refusals—those grounded in safety or appropriateness concerns—were more strongly penalized than technical refusals that signal the model's limitations. Although both refusal types led to a decline in user preference, the drop was significantly steeper for ethical refusals. These findings suggest that technical refusals are seen as more understandable or acceptable, while ethical refusals may be perceived as overly cautious, evasive, or even patronizing—particularly when the model offers no opportunity for clarification or dialogue.

Taken together, these findings underscore a core tension in LLM deployment: the very behaviors that signal strong alignment—such as declining ethically risky prompts—can simultaneously undermine user satisfaction. This highlights a deeper challenge in alignment research, where users and developers may hold differing expectations of what constitutes "helpful" or "trustworthy" AI behavior. Importantly, we do not take a normative stance on how often models *should* refuse; rather, we aim to empirically document how such refusals are actually perceived in practice by users.

### 5.2. Refusal Design Strategy: Mitigating the Penalty (H3 & H4)

Despite the robust refusal penalty identified in our analysis—especially for ethical refusals—our findings also point to specific strategies that can help mitigate this dissatisfaction. Given that safety alignment remains a crucial design objective, the solution may not be to eliminate refusals entirely, but to optimize *how* and *when* they are delivered. Drawing on H3 and H4, we identify three actionable insights for designing more effective refusal strategies:

**1. Avoid overly formulaic refusals.**
Refusals that rely on generic templates—such as "*I am sorry I cannot help with that.*"—were consistently associated with lower win rates. These responses often fail to engage with the user's prompt and are perceived as unhelpful or dismissive. H3a highlights that refusals with *higher semantic similarity* to the user prompt are judged more favorably, suggesting that tailoring refusal language to reflect the user's intent improves perceived responsiveness.

**2. Provide context, reasoning, or alternatives.**
Similarly, longer responses that include explanations or offer constructive alternatives reduce dissatisfaction. As suggested by H3b, refusal length was positively associated with win rates for ethical refusals, indicating that users appreciate when the model "makes an effort" rather than shutting down the conversation. In contrast, for technical refusals, text length did not significantly improve ratings.

**3. Calibrate moderation to avoid unnecessary refusals.**
Our analysis of prompt sensitivity (H4) shows that users are more accepting of ethical refusals when the underlying prompt is clearly inappropriate or harmful (e.g., involving hate speech or illegal activities). However, the majority of ethical refusals were triggered by prompts not flagged by the moderation system—including lighthearted, hypothetical, or ambiguous queries. These potentially overcautious refusals were especially penalized, suggesting that over-moderation harms user satisfaction without necessarily improving safety outcomes. Improving moderation calibration—so that refusals better align with genuine content risks—can therefore enhance both trust and usability. Recent work such as SORRY-Bench (Xie et al., 2024) has begun to explore calibration techniques and evaluation strategies for refusals, but this remains an open area of research.

That said, interpretations of specific content categories should be made with caution. For example, we find that prompts flagged as sexual or violent content are associated with particularly low user win rates for refusal responses, possibly indicating that users perceive these refusals as overly restrictive or miscalibrated. However, such reactions are likely to vary depending on cultural norms, platform context, or user demographics. What one group considers appropriate or benign, another might find objectionable. Moreover, in contexts where platforms are accessible to minors, stricter moderation of such content may reflect legitimate safeguarding concerns (Kurian, 2024). These findings underscore the importance of audience-aware moderation strategies that balance user satisfaction with protective responsibilities.

### 5.3. Limitations and Future Research

While this study offers novel insights into refusal behavior in LLMs, several limitations should be acknowledged.

First, our classification scheme relies on five relatively broad categories (normal and ethical/technical refusals and disclaimers). While these distinctions are conceptually grounded and empirically robust, they may oversimplify the nuanced motivations behind LLM responses. Future work could refine the taxonomy further—for instance, by distinguishing between legal, moral, and reputational concerns within ethical refusals, or by incorporating additional dimensions such as whether the model provides an explicit explanation for its refusal, offers alternatives, or conveys uncertainty.

Second, our analysis is based on data from the Chatbot Arena, a valuable but self-selected benchmark. Arena participants may not represent the broader population of end-users across domains, cultures, or languages. As such, generalizations about user satisfaction should be made with care. Complementary studies using targeted user surveys or in-domain deployment data could strengthen the generalizability of our findings.

Third, our use of the OpenAI Moderation API as a proxy for content sensitivity—while practical and grounded in prior work—reflects one interpretation of policy-relevant content. The moderation categories are not exhaustive, and the binary flagging mechanism may miss subtler cases of inappropriate or controversial input. Future research might explore multi-dimensional models of sensitivity, incorporating user-specific norms or platform-specific guidelines.

Fourth, while our win/loss outcomes offer a behavioral proxy for satisfaction, they do not capture why users prefer one response over another. Interpretations are necessarily indirect. Mixed-method approaches, including qualitative feedback or think-aloud protocols, could provide richer insight into the perceived helpfulness, tone, and appropriateness of refusal responses.

Finally, our analysis of refusal phrasing relies on two relatively coarse proxies: textual similarity and response length. While these features provide a useful starting point—demonstrating that more aligned and detailed refusals tend to be better received—they do not capture the full range of communicative nuance. Future research could investigate more fine-grained linguistic and rhetorical features, such as tone, politeness strategies, hedging, or explanatory structure. A deeper understanding of these stylistic dimensions may offer more targeted guidance for refusal generation and help bridge the gap between alignment and user experience.

## 6. Conclusion

This study investigated how different types and phrasings of refusal behavior in large language models affect user satisfaction. Grounded in a novel refusal typology and a large-scale, annotated dataset, we tested four hypotheses spanning user perceptions, response phrasing, and contextual moderation effects.

We found consistent support for H1: refusal responses—especially outright ethical refusals—significantly reduce user satisfaction in pairwise comparisons. Our results also supported H2a, showing that ethical refusals are more negatively received than technical ones. In contrast, H2b, which proposed the opposite relationship under certain conditions, was not supported.

In line with H3, we showed that refusal phrasing matters: longer and more contextually aligned refusals were judged more favorably by users. While our measures were coarse, they offer a first proxy for understanding how refusal strategy affects user reception. Lastly, our analysis of prompt sensitivity (H4) confirmed that refusals are penalized less when issued in response to flagged or clearly inappropriate content—emphasizing the importance of accurate moderation and calibration.

Taken together, these findings reveal both challenges and design opportunities in the deployment of aligned LLMs. While refusals are often necessary for safety and compliance, their social perception depends not only on what is refused, but how and why. Future work should build on these distinctions to refine both refusal design and evaluation frameworks.

# Appendix

**Table AT1: OLS Regressions – LLM-Refusals and Win-Rates (incl. Opp. Model Coefficients)**

| | User Decision | | |
|---|---|---|---|
| Variable | Win | Loss | Tie |
| Disclaimer Ethical | -0.073*** | 0.113*** | -0.040** |
| | (0.019) | (0.018) | (0.019) |
| Disclaimer Technical | -0.098*** | 0.106*** | -0.008 |
| | (0.009) | (0.009) | (0.009) |
| Refusal Ethical | -0.375*** | 0.320*** | 0.055*** |
| | (0.011) | (0.013) | (0.014) |
| Refusal Technical | -0.214*** | 0.162*** | 0.052*** |
| | (0.011) | (0.012) | (0.013) |
| Disclaimer Ethical Opp. | 0.139*** | -0.092*** | -0.047** |
| | (0.018) | (0.017) | (0.019) |
| Disclaimer Technical Opp. | 0.112*** | -0.108*** | -0.004 |
| | (0.009) | (0.009) | (0.009) |
| Refusal Ethical Opp. | 0.305*** | -0.370*** | 0.064*** |
| | (0.013) | (0.011) | (0.015) |
| Refusal Technical Opp. | 0.179*** | -0.214*** | 0.034*** |
| | (0.013) | (0.010) | (0.013) |
| Similarity | 0.006** | -0.011*** | 0.005* |
| | (0.003) | (0.003) | (0.003) |
| Similarity Opp. | -0.012*** | 0.013*** | -0.001 |
| | (0.003) | (0.003) | (0.003) |
| Length | 0.079*** | -0.058*** | -0.020*** |
| | (0.003) | (0.003) | (0.003) |
| Length Opp. | -0.055*** | 0.078*** | -0.023*** |
| | (0.003) | (0.004) | (0.003) |
| Controls LLM Model | Yes | Yes | Yes |
| Observations | 49938 | 49938 | 49938 |
| R-squared | 0.13 | 0.13 | 0.02 |

Robust standard errors in parenthesis.* p<.1, ** p<.05, ***p<.01. Controls LLM includes dummy variables for the LLM used for the response.